\documentclass{article}

\usepackage{microtype}
\usepackage{graphicx}
\usepackage{subcaption}
\usepackage{booktabs} 

\usepackage{hyperref} 
\usepackage[preprint]{icml2026}
\usepackage{xcolor}
\usepackage{listings}
\usepackage{makecell}

\usepackage{amsmath,amsfonts,bm}

\def\eqref#1{equation~\ref{#1}}

\def\1{\bm{1}}

\DeclareMathAlphabet{\mathsfit}{\encodingdefault}{\sfdefault}{m}{sl}
\SetMathAlphabet{\mathsfit}{bold}{\encodingdefault}{\sfdefault}{bx}{n}

\usepackage{soul}

\usepackage{amsmath}
\usepackage{amssymb}
\usepackage{mathtools}
\usepackage{amsthm}     
\usepackage{url}
\usepackage{multirow}
\usepackage{caption}
\usepackage{booktabs}
\usepackage{xspace}
\usepackage{tabularx}
\usepackage{array}
\usepackage{siunitx}
\usepackage{collcell}
\usepackage[table]{xcolor}
\usepackage{tcolorbox}

\icmltitlerunning{Don't Claim Benchmark-Oriented Optimization Improves General Coding Capability — Diverse Evaluation Is Required}

\begin{document}

\twocolumn[
  \icmltitle{Don't Claim Benchmark-Oriented Optimization Improves General Coding Capability — Diverse Evaluation Is Required}

  \icmlsetsymbol{equal}{*}

  \begin{icmlauthorlist}
    \icmlauthor{Egor Shibaev}{equal,de,unide}
    \icmlauthor{Vera Kudrevskaia}{equal,de,unide}
    \icmlauthor{Timur Galimzyanov}{equal,de}
    \icmlauthor{Mikhail Evtikhiev}{equal,cy}
    \icmlauthor{Ana Terna}{nl}
    \icmlauthor{Rastislav Rabatin}{uk}
    \icmlauthor{Timur Kudashev}{de,unide}
    \icmlauthor{Timofey Bryksin}{cy}
    \icmlauthor{Arina Puchkova}{de}
    \icmlauthor{Patrik Bartak}{nl}
    \icmlauthor{Egor Bogomolov}{nl}
    \icmlauthor{Sergey Titov}{nl}
  \end{icmlauthorlist}

  \icmlaffiliation{de}{Code Modelling Research, JetBrains Research, Munich, Germany}
  \icmlaffiliation{nl}{Code Modelling Research, JetBrains Research, Amsterdam, The Netherlands}
  \icmlaffiliation{uk}{Code Modelling Research, JetBrains Research, London, United Kingdom}
  \icmlaffiliation{cy}{Code Modelling Research, JetBrains Research, Paphos, Cyprus}
  \icmlaffiliation{unide}{Constructor University, Bremen, Germany}
  
  \icmlcorrespondingauthor{Mikhail Evtikhiev}{mikhail.evtikhiev@jetbrains.com}

  \icmlkeywords{LLM, benchmarks, evaluation, machine learning for software engineering}

  \vskip 0.3in
]
                   
\printAffiliationsAndNotice{} 
\begin{abstract}
Post-training papers, model cards, and blog posts often treat scores on a small set of coding benchmarks (e.g., SWE-bench and LiveCodeBench) as evidence of broad ``coding capability'', both for research artifacts and  user-facing systems. 
\textbf{We argue that optimization for these benchmarks leads to measuring task-specific performance, creating a \emph{meaning gap} between measured scores and claims of general coding ability.}
We examine this gap with a  Django-based case study benchmark suite we create.

Evaluating foundation models and checkpoints post-trained on SWE-bench trajectories, we find that benchmark rankings frequently fail to generalize. 
Post-trained checkpoints show little cross-task transfer, and SWE-bench optimization yields limited or no gains on our tasks or on LiveCodeBench. 
Similarly, fine-tuning on individual Django modalities fails to transfer.

We conclude that a small number of benchmarks is insufficient for evaluating diverse models under benchmark optimization pressure. 
We encourage the community to use differentiated evaluation—holistic assessment for frontier models, multi-task suites for research, and human-in-the-loop studies for narrow task applications. 
Finally, we argue for creating a capability taxonomy and sustained benchmark maintenance, rather than one-off benchmark releases.
Without reliable evaluation standards, engineers and researchers using LLMs and agents have to rely on insufficient evidence to make research, development, and deployment decisions.

\end{abstract}

\section{Introduction}
The deep learning for code (DL-for-code) community has converged on a narrow evaluation paradigm.
On one end, researchers use self-contained algorithmic tasks, such as HumanEval, that are convenient for benchmarking, yet only weakly resemble real-world software development. 
On the other end, SWE-bench~\cite{Jimenez2023SWEbench} became the de facto standard for measuring ``real-world'' coding ability, with leaderboard rankings frequently interpreted as proxies for general coding capability.
``General coding capability'' is a latent factor explaining positive correlations in model performance across diverse programming tasks, distinct from narrow task-specific skills.
It is similar to the definition of intelligence by Chollet~\cite{chollet2019measure}.
This convergence has driven a remarkable engineering effort: complex post-training approaches, specialized agent architectures, and training pipelines specifically designed to maximize SWE-bench scores~\cite{zeng2025skywork, pan2024training}.

SWE-bench scores may indeed correlate with genuine progress in coding.
Recent generations of foundation models show clear improvements across many coding-related behaviors, and agents combined with frontier proprietary models are now widely used in practice~\cite{aitw2025dashboard}. 
However, the current structure of SWE-bench and other coding benchmarks often cannot tell us \emph{why} scores improve.
SWE-bench-style performance mixes multiple factors: repository understanding, patch synthesis, tool use, search and retrieval strategies, and adherence to a particular workflow and output format. 
It is thus difficult to identify the true driver of improvement without deep analysis. 
For the foundation models, improvements are typically reported across a broad range of benchmarks, including non-coding ones, providing stronger evidence of genuine gains in general intelligence. 
In contrast, many post-training papers report improvements primarily on SWE-bench (or a small cluster of similar benchmarks), providing a potentially limited and ambiguous signal about generalization and changes in underlying capabilities.

This conflation matters because it shapes research priorities. 
For example, if post-training on SWE-bench trajectories reliably improves general coding capabilities, this approach provides a path towards model improvement. 
But if it just produces models skilled at SWE-bench-like tasks, the field risks optimizing for the benchmark itself rather than addressing the underlying capability or other coding tasks, severely limiting the impact of this post-training approach.
Thus, researchers working on evaluation of coding agents and LLM coding capabilities may be confounded into misinterpreting their results, and engineers relying on their work may make suboptimal deployment decisions.

Our evidence supports the latter interpretation: post-training gains on SWE-bench do not consistently transfer to other code tasks, even within the same repository.
To illustrate this, we build a Django benchmark covering code editing, generation, and completion, and evaluate community-released checkpoints and our fine-tuned models.
We observe a consistent pattern across both community-released SWE-bench-oriented checkpoints and models we fine-tune.
Models fine-tuned on issue-resolution trajectories do not generally improve on our Django-related tasks or LiveCodeBench (LCB)~\cite{jain2024livecodebench}, while models fine-tuned on one of the Django tasks do not improve on other Django tasks or LCB.
This suggests narrow training recipes drive task-specific specialization, not general coding improvement.
Moreover, SWE-bench (and even multi-benchmark) rankings may not always reliably predict relative performance on these modalities. 
Together, our findings suggest that SWE-bench rankings may systematically misrepresent models' relative strengths on individual SE tasks, particularly since real-world usage extends beyond agentic issue resolution.

\textbf{In this position paper, we argue that this mismatch between performance on different benchmarks reflects a \emph{construct validity} problem: when a small number of benchmarks is treated as a proxy for ``general coding capability,'' the resulting claims exceed what the measurements can actually support}.
SWE-bench targets a complex but specific behavior that involves navigating repositories, understanding issue descriptions, and producing targeted patches. 
These measurements may mix multiple underlying capabilities with narrow task-specific skills. 
High performance may indicate strong coding ability, but may equally reflect proficiency at the particular format and workflow developed under benchmark optimization pressure. 
Without evaluation across diverse task modalities, we cannot distinguish among these hypotheses, understand the strengths and weaknesses of our post-training approaches, or even assess the foundation models with sufficient granularity.
The resulting misnterpretation can then spread to engineers and users working with LLMs and agents, as LLM-assisted coding and coding agents become increasingly common, and, in the end, undermine their trust in these systems.

We call for a fundamental shift in how the community evaluates code models. 
Over-reliance on a narrow set of benchmarks has led the community to equate task performance with general coding capability, undermining construct validity and making it impossible to distinguish genuine ability from benchmark-specific optimization. 
Evaluation frameworks should explicitly test for transfer across task modalities, separate general coding improvement from narrow specialization, and provide a complete picture of model capabilities and limitations.
Our benchmark suite and experimental methodology illustrate a possible path forward, but our main recommendation is methodological: the field must move beyond leaderboard rankings towards evaluation practices that more directly measure the constructs we claim to care about.
As a practical step in this direction, we propose a three-pronged approach: holistic assessment for frontier models, diverse multi-task suites for incremental research and smaller models, and human-in-the-loop assessment for narrow applications.
\section{Background}
\subsection{Coding benchmarks}

The evaluation of code-oriented LLMs broadly falls into two families.

\textbf{Self-contained code tasks} package everything needed to solve a task within the prompt itself. 
Many widely used suites in this family focus on short-form code generation. 
A typical example is HumanEval~\cite{Chen2021Codex} that contains 164 hand-crafted Python problems for function-level generation.
Performance is typically reported with pass@k (often pass@1), \textit{i.e.}, whether a sampled solution passes the provided tests.
For example of a broader benchmark suite, LiveCodeBench~\cite{jain2024livecodebench} adds self-repair, code execution, and test output prediction to code generation.

\textbf{Repository-level} benchmarks evaluate a model within an entire codebase.
Given a repository snapshot and a natural-language task description, the model must produce a patch that resolves the task and passes the tests. 
SWE-bench Verified~\cite{chowdhury2024swebenchverified} is the \textit{de facto} standard in this family, comprising 500 issue resolution instances drawn from GitHub issues across 12 popular Python repositories. 
While issue resolution is an important task, it does not represent the whole space of coding agent capabilities.
As of late, SWE-bench Pro~\cite{deng2025swe}, which also targets issue resolution tasks, also started gaining traction.
Outside of issue resolution, the only relatively popular benchmark is TerminalBench~\cite{merrill2026terminal}.

The evaluation approaches of these families measure different capabilities, and it is unclear whether the benchmark scores should always be correlated. 
Strong performance on self-contained tasks may fail to predict success on repository-level tasks that require navigating codebases, understanding dependencies, and producing targeted edits. 
It is moreover unclear, whether these two families of benchmarks cover general coding capability.

\subsection{Foundation models}
Modern foundation models are trained for broad, multi-domain competence, and model cards and technical reports summarize coding performance with a small set of representative benchmarks. 
The ability of model creators to benchmark these models for a particular domain is naturally limited by time and the necessity to produce concise, readable reports. 
For example, the GPT-5 model card includes reports on only two coding benchmarks (SWE-bench Verified and Aider Polyglot), and the Qwen3 technical report~\cite{yang2025qwen3} includes four benchmarks, of which three belong to the self-contained algorithmic tasks, and the fourth is execution-reasoning-oriented CRUXEval.
The limited number of reported benchmarks, combined with the possibility of little correlation across benchmarking tasks, implies the risk that some coding capabilities may fall into a blind spot of benchmark suites used by the foundation model creators, resulting in a misleading evaluation.

\subsection{Post-training foundation models}
These limitations become even more pronounced for post-training checkpoints. 
Compared to releases of foundation models, post-training papers face page limits, computational costs of evaluation~\cite{jordan2024position}, and the engineering overhead of adopting additional harnesses. 
As a result, single-benchmark evaluations are often the pragmatic default rather than an exception.

This constraint produces a typical pattern: post-training works tend to evaluate on benchmarks aligned with their intended contribution.
Works targeting repository-level SE commonly report only SWE-bench performance (e.g., Lingma SWE-GPT~\cite{ma2024lingma}, R2EGym-Agent~\cite{jain2025r2e}, SWE-agent-LM~\cite{yang2025swe}, Skywork-SWE~\cite{zeng2025skywork}, DeepSWE-Preview~\cite{deepswe2025}). 
On the other hand, studies focused on developing general post-training techniques often evaluate on self-contained suites like HumanEval (e.g., ~\cite{wei2024selfcodealign}, \cite{tang2025optimizing}, \cite{yu2024reasoning}).
While perfectly rational, this evaluation approach does not allow distinguishing whether the improvement comes from task-specific optimization or reflects a general improvement of coding capability.
Capability generalization cannot be taken for granted: Yu et al.~\cite{yu2024humaneval} show cases where instruction tuning improves algorithmic benchmark scores without commensurate gains on the underlying behavior.
This cited pattern we find undesirable motivates our position. 
To support it, we directly test cross-task transfer by evaluating checkpoints trained on specific tasks across multiple SE modalities.

\section{Benchmark measurements and capability claims}

From the general coding capability point of view, any single benchmark can at best provide a noisy, partial measurement of this latent capability. 
In particular, when two models are far apart in overall competence (e.g., frontier systems versus small baselines), a large and consistent performance gap on a benchmark is often a reasonable shorthand for a capability difference. 
The difficulty arises when benchmark results are treated as sufficient evidence for broad capability claims, especially when models are optimized for that benchmark.

Reporting model performance on a limited number of benchmarks is practical and reasonable. 
However, there seems to be a systematic disconnect between what code benchmarks actually measure and how benchmark scores are commonly interpreted.
Because of this ``meaning gap'', specific achievements on a narrow benchmark may inflate into broad assertions of general coding capability.

We illustrate the meaning gap concept using examples from the Qwen3-Coder-480B-A35B model descriptions.
The original blog post reports the model performance on eight coding-related tasks, five of which correspond to issue resolution similar to SWE-bench~\cite{qwen3ai}, and summarizes them as ``exceptional performance in both coding and agentic tasks''.
Finally, an independent blog post comparing models makes a claim ``Chinese models aren't just competing---they're winning. Qwen 3 Coder leads at 67\% on SWE-bench, surpassing GPT-4.1's 54.6\%''~\cite{chinesegpt4}. 
Each step is understandable in isolation, but together they broaden a narrow performance measurement into a claim about general superiority at coding that may influence deployment decisions.

Conceptually, the meaning gap combines two processes: (i) generalization from benchmark scores to capability statements by the model and checkpoint creators, and (ii) further amplification of statements from scientific papers and technical reports by the general audience.
While the latter problem may have a strong impact on the DL-for-code community, it merits a separate sociological study beyond the scope of our paper.
Thus, in our study, we focus on the relationship between benchmark scores and the general coding capability that the community seeks to improve, and use the ``meaning gap'' term to refer to the mismatch between the two.

The meaning gap is directly related to the construct validity problem.
In their review on LLM benchmarks, Bean et al.~\cite{bean2025measuring} provide recommendations for benchmark descriptions that would strengthen their construct validity.
Describing SWE-bench Verified according to their checklist yields the following description. 
\begin{tcolorbox}[colback=yellow!20, colframe=black, width=\columnwidth, boxsep=2mm]
\textit{SWE-bench is an issue resolution capability benchmark.
It is based on 12 Python repositories with a highly skewed task distribution, with 46\% of 500 examples coming from the Django project.
SWE-bench Verified does not address possible data contamination issues, and the common practice of reporting its scores does not include uncertainty estimates or analysis of common failure modes.}
\end{tcolorbox}

These observations do not diminish the value of SWE-bench Verified as a benchmark.
However, when a model achieves a high score on SWE-bench Verified, there is no standard way to determine whether the improvement reflects (a) genuine SE capability that transfers to other coding tasks, (b) limited SE capability on this family of tasks, (c) inadvertent optimization for SWE-bench's specific format, tasks, workflow, or agentic scaffolding, (d) accidental contamination of training dataset, or (e) some mixture of all of the above.
This problem is especially pronounced in academic post-training studies, as researchers often cannot afford to run many benchmarks and may mistake within-benchmark gains for general improvements in capability.

At this point, we only identify the meaning gap and argue that it can be practically consequential. 
To test it, we create a benchmark suite that measures model performance across multiple in-repository task modalities, explicitly assessing the cross-task transfer capability.
If models with higher SWE-bench Verified scores perform worse on our suite, then SWE-bench Verified is a weak predictor for these tasks and is insufficient as a standalone proxy for broader coding capability.
In the next section, we describe our benchmark and evaluate a range of models and checkpoints on it.

\section{Evidence from a Django case study}
\label{sec:experiments}
\subsection{Benchmarks and data}


To study the meaning gap in a controlled setting, we build a benchmark suite over a single real-world repository, spanning three common SE modalities: (i) method generation, (ii) method completion, and (iii) program repair. 
We use these experiments to test plausibility of our position.
We describe all benchmark details in the appendices. 

\textbf{Repository snapshot.} We choose Django because it constitutes 46\% of SWE-bench Verified. 
All tasks are derived from a snapshot at version 4.0.4.\footnote{commit hash $89807f bde8b7b17d00434bc4695535855e96fe77$, date: 11th Apr 2022}.

This choice enables a simple and testable prediction: if SWE-bench-optimized checkpoints truly improve transferable repository-level SE ability (rather than specializing to SWE-bench’s workflow and distribution), then they should also improve on Django-centric tasks within the same codebase.
Thus, we test cross-task transfer for relatively close tasks that come from data distribution close to SWE-bench's.
We describe repository selection, data statistics, and data curation in appendix~\ref{app:repo}.


\textbf{Docstring standardization.} Django's in-code documentation has uneven quality, which can make ``generate the method from its docstring'' task instances noisy and inconsistently specified. 
To reduce this variance, we synthesize standardized docstrings for all methods with DeepSeek-R1~\cite{guo2025deepseek} and merge them with the original comments whenever applicable.
We provide extensive description of our documentation synthesis pipeline in Appendix~\ref{app:docs}.

\textbf{Task definitions.} Using this augmented data, we build three benchmarks: 
\begin{itemize}
    \item \textbf{Method generation:} the model receives a target method signature, its docstring, and the source file with the target method body removed (method declaration is provided); it must generate the missing body.
    \item \textbf{Method completion:} the model receives the preceding file content with the first half of the target method; it must generate the remaining half (in the chat format). 
    \item \textbf{Program repair:} the model receives a source file in which one method has been corrupted, and a stack trace produced by running the Django test suite on the repo with this broken file. 
    The model must identify the broken method and output the correct version.
\end{itemize}
We run our benchmarks without agentic scaffolding to check in isolation how model post-training affects model coding behavior. 
Appendix~\ref{app:scaffolding} gives one scaffolding comparison to illustrate how scaffolding can affect performance without resolving the cross-task transfer concern. 

\textbf{Train/test split and leakage control.} Each benchmark is paired with a corresponding instruction-tuning dataset constructed from the same task template.
After splitting, the dataset contains 3,180 train samples and 359 test samples.
To avoid cross-contamination through shared class hierarchies, we split at the level of inheritance trees: all ancestor and descendant classes of each method's defining class go to the same split. 
We do not train on Django's test suite, and we do not include tests in evaluation prompts.
We describe train/test split in appendix~\ref{app:split}, and provide prompts in appendix~\ref{app:prompts}.

\textbf{Evaluation.}
We evaluate candidate outputs by running the official Django 4.0.4 test suite and report \textbf{pass@1}. For all benchmarks, we use greedy decoding at $T=0$, so each instance has a single attempt.

\textbf{Out-of-suite comparison: LiveCodeBench.}
In addition to our benchmarks, we evaluate all models and checkpoints on LiveCodeBench (LCB), using tasks added after October 1, 2024 (341 examples).
We use LCB to validate basic code generation capabilities on self-contained algorithmic problems, contrasting with our Django benchmarks that assess repository-level understanding and task execution, and testing for the more distant cross-task transfer.
We again evaluate with greedy decoding at $T=0$.

\subsection{Models, checkpoints and fine-tuning}
We evaluate three groups: (i) foundation models, (ii) public SWE-bench-oriented checkpoints, and (iii) our task-specific fine-tunes.

\textbf{Foundation models.}
We include Qwen3-32B~\cite{hui2024qwen2}, Qwen2.5-Coder-7B-Instruct, and Qwen2.5-Coder-32B-Instruct~\cite{yang2025qwen3} as the representative strong open models spanning sizes and releases, and the popular base models for public checkpoints.

\textbf{Public checkpoints}. We evaluate public checkpoints obtained by fine-tuning models on agentic issue-resolution trajectories (e.g., SWE-bench-style traces): DeepSWE-Preview~\cite{deepswe2025}, R2EGym-7B-Agent~\cite{jain2025r2e}, SWE-agent-LM-7B~\cite{yang2025swe}, Openhands LM 32B~\cite{openhands}, Skywork-SWE~\cite{zeng2025skywork}, SWE-agent-LM-32B, R2EGym-32B-Agent. 
Most of these checkpoints are based on the Qwen2.5-Coder-Instruct model. 
These checkpoints are designed to improve SWE-bench performance. 
By testing them on our Django benchmarks, we study whether improvements transfer to other repository-level tasks in the same codebase—expected if coding capabilities genuinely improved.
LCB tests for transfer to more distant, self-contained algorithmic tasks.

\textbf{Our fine-tuning experiments.} To isolate the role of \emph{task-specific} supervision, we fine-tune Qwen2.5-Coder-7B-Instruct and Qwen2.5-Coder-32B-Instruct on each of the three task-specific datasets (generation, completion, repair).
We choose these models because most of the public checkpoints we evaluate were built on top of one of them.
In every case, we use LoRA with learning rate $1e-5$, and use the last epoch checkpoint for evaluation; we provide full training details in appendix~\ref{app:finetuning}.
We \textbf{do not aim} to exhaustively optimize these models. 
The goal is to test a practical, ``reasonable'' fine-tuning recipe and measure \emph{cross-task transfer} rather than to maximize within-task scores.

\subsection{Results}

\sisetup{
table-number-alignment = center,
  round-mode = places,
  round-precision = 1,
}

\begin{table*}[!t]

\caption{Cross-task transfer diagnostics for SWE-bench-oriented checkpoints}
\label{tab:model_comparison}
\begin{tabularx}{\textwidth}{
    @{}
    >{\raggedright\arraybackslash}p{0.11\textwidth}
    >{\raggedright\arraybackslash}p{0.18\textwidth}
    *{10}{@{\hspace{5pt}}S[table-format=3.1]@{\hspace{5pt}}}
    @{}
}
\toprule
 &  & \multicolumn{2}{c}{SWE-bench} & \multicolumn{2}{c}{Generation} & \multicolumn{2}{c}{Completion} & \multicolumn{2}{c}{Repair} & \multicolumn{2}{c}{LCB} \\
 &  & {score} & {$\Delta$} & {score} & {$\Delta$} & {score} & {$\Delta$} & {score} & {$\Delta$}& {score} & {$\Delta$}\\
\midrule
\multirow{5}{=}{\parbox[t]{\hsize}{Qwen2.5-Coder-32B}} & Base model & 7.00 & \cellcolor[HTML]{E6E6E6} {--} & 64.07 & \cellcolor[HTML]{E6E6E6} {--} & 60.17 & \cellcolor[HTML]{E6E6E6} {--} & 47.35 & \cellcolor[HTML]{E6E6E6} {--} & 28.2 & \cellcolor[HTML]{E6E6E6} {--} \\
 & Openhands LM 32B & 37.20 & {\cellcolor[HTML]{7CB17C}} 30.20 & 66.57 & {\cellcolor[HTML]{AECFAE}} 2.51 & 48.75 & {\cellcolor[HTML]{D9A0A0}} -11.42 & 13.93 & {\cellcolor[HTML]{D9A0A0}} -33.43 & 25.8 & {\cellcolor[HTML]{E5BBBB}} -2.3 \\
 & Skywork-SWE & 38.00 & {\cellcolor[HTML]{79AF79}} 31.00 & 68.80 & {\cellcolor[HTML]{71AA71}} 4.74 & 57.66 & {\cellcolor[HTML]{F5DFDF}} -2.51 & 41.78 & {\cellcolor[HTML]{F7E3E3}} -5.57 & 27.3 & {\cellcolor[HTML]{F1D6D6}} -0.9 \\
 & SWE-agent-LM-32B & 40.20 & {\cellcolor[HTML]{71AA71}} 33.20 & 66.57 & {\cellcolor[HTML]{AECFAE}} 2.51 & 55.15 & {\cellcolor[HTML]{EDCDCD}} -5.01 & 40.11 & {\cellcolor[HTML]{F5DFDF}} -7.24 & 25.2 & {\cellcolor[HTML]{D9A0A0}} -2.9 \\
  & R2EGym-32B-Agent & 34.40 & {\cellcolor[HTML]{87B787}} 27.40 & 67.41 & {\cellcolor[HTML]{97C197}} 3.34 & 58.22 & {\cellcolor[HTML]{F6E3E3}} -1.95 & 40.67 & {\cellcolor[HTML]{F5E0E0}} -6.69 & 29.6 & {\cellcolor[HTML]{71AA71}} 1.5  \\
\midrule
\multirow{3}{=}{\parbox[t]{\hsize}{Qwen2.5-Coder-7B}} & Base model & 1.80 & \cellcolor[HTML]{E6E6E6} {--} & 52.92 & \cellcolor[HTML]{E6E6E6} {--} & 50.97 & \cellcolor[HTML]{E6E6E6} {--} & 17.83 & \cellcolor[HTML]{E6E6E6} {--} & 15.2 & \cellcolor[HTML]{E6E6E6} {--} \\
 & R2EGym-7B-Agent & 19.00 & {\cellcolor[HTML]{71AA71}} 17.20 & 54.87 & {\cellcolor[HTML]{71AA71}} 1.95 & 41.23 & {\cellcolor[HTML]{F5E0E0}} -9.75 & 15.04 & {\cellcolor[HTML]{F6E3E3}} -2.79 & 17.3 & {\cellcolor[HTML]{71AA71}} 2.1 \\
 & SWE-agent-LM-7B & 15.20 & {\cellcolor[HTML]{8DBB8D}} 13.40 & 5.85 & {\cellcolor[HTML]{D9A0A0}} -47.08 & 2.23 & {\cellcolor[HTML]{D9A0A0}} -48.75 & 1.67 & {\cellcolor[HTML]{D9A0A0}} -16.16 & 6.7 & {\cellcolor[HTML]{D9A0A0}} -8.5 \\
\midrule
\multirow{2}{=}{\parbox[t]{\hsize}{Qwen3-32B}} & Base model & 23.00 & \cellcolor[HTML]{E6E6E6} {--} & 67.41 & \cellcolor[HTML]{E6E6E6} {--} & 43.45 & \cellcolor[HTML]{E6E6E6} {--} & 46.80 & \cellcolor[HTML]{E6E6E6} {--} & 61.00 & \cellcolor[HTML]{E6E6E6} {--} \\
 & DeepSWE-Preview & 42.20 & {\cellcolor[HTML]{71AA71}} 19.20 & 67.69 & {\cellcolor[HTML]{71AA71}} 0.28 & 44.29 & {\cellcolor[HTML]{71AA71}} 0.84 & 51.25 & {\cellcolor[HTML]{71AA71}} 4.46 & 60.1 & {\cellcolor[HTML]{D9A0A0}} -0.9 \\
\bottomrule
\end{tabularx}
\end{table*}


Table~\ref{tab:model_comparison} summarizes the performance of foundation models and public SWE-bench-oriented checkpoints on our Django suite and LiveCodeBench (with deltas computed against the corresponding base model). 
For each checkpoint, we show the absolute score and its change ($\Delta$) relative to the corresponding base model. 
We take SWE-bench Verified scores from the original papers and reports.
To assess whether fine-tuning for SWE-bench transfers to out-of-domain tasks, we examine the pattern of performance changes.
We use these results to diagnose the problem and illustrate failure of construct validity and risks of cross-task transfer. 
We do not interpret them as a new leaderboard.

\textbf{Foundation models: rankings are not always stable across modalities.}
Even for foundation models, relative ordering can flip across tasks: a model stronger on SWE-bench can underperform on our benchmark, illustrating that a single headline score may not reflect a full capability profile. 
For example, while Qwen-2.5-32B-Coder-Instruct outperforms Qwen3-32B on our code completion benchmark, it has worse SWE-bench scores.

\textbf{Public checkpoints: no cross-task transfer.} SWE-bench-optimized checkpoints, while significantly outperforming their base models on SWE-bench, generally do not improve on our Django tasks or LiveCodeBench. 
Across the 28 out-of-domain checkpoint--benchmark comparisons (7 checkpoints $\times$ 4 benchmarks), we observe 18 degradations versus ten improvements. 
Five of seven checkpoints degrade on the majority of benchmarks, none improve on all four, and only DeepSWE-Preview shows more improvements than degradations overall. 
Because we report single greedy evaluations on stochastic benchmarks, small deltas should not be over-interpreted. 
The overall pattern is consistent with the absence of change in cross-task capabilities, with occasional regressions. 
Importantly, these regressions do not always come from output-format failures: most checkpoints follow task formats, with only SWE-agent-LM-7B and OpenHands-LM-32B sometimes failing to follow instructions.
To illustrate our point on cross-task transfer for both format and substance mistakes, we provide error-flow analysis in appendix~\ref{app:error-flow}. 

\begin{table*}[!t]
\caption{Cross-task transfer diagnostics for task-specific fine-tuning}
\label{tab:model_comparison_finetune}
\begin{tabularx}{\textwidth}{
    @{}
    >{\raggedright\arraybackslash}p{0.18\textwidth}
    >{\raggedright\arraybackslash}p{0.18\textwidth}
    *{8}{S[table-format=3.1]}
    @{}
}
\toprule
 &  & \multicolumn{2}{c}{Generation} & \multicolumn{2}{c}{Completion} & \multicolumn{2}{c}{Repair} & \multicolumn{2}{c}{LCB} \\
 & {Fine-tuning} & {score} & {$\Delta$} & {score} & {$\Delta$} & {score} & {$\Delta$} & {score} & {$\Delta$} \\
\midrule
\multirow{4}{=}{\parbox[t]{\hsize}{Qwen2.5-Coder-32B}} & Base Model & 64.07 & \cellcolor[HTML]{E6E6E6} {--} & 60.17 & \cellcolor[HTML]{E6E6E6} {--} & 47.35 & \cellcolor[HTML]{E6E6E6} {--} & 28.20 & \cellcolor[HTML]{E6E6E6} {--} \\
& Generation & 71.03 & {\cellcolor[HTML]{71AA71}} 6.96 & 54.60 & {\cellcolor[HTML]{D9A0A0}} -5.57 & 42.90 & {\cellcolor[HTML]{F5E1E1}} -4.45 & 22.90 & {\cellcolor[HTML]{D9A0A0}} -5.30 \\
 & Completion & 66.02 & {\cellcolor[HTML]{CEE2CE}} 1.95 & 74.65 & {\cellcolor[HTML]{71AA71}} 14.48 & 24.51 & {\cellcolor[HTML]{D9A0A0}} -22.84 & 25.20 & {\cellcolor[HTML]{E9C4C4}} -2.90 \\
 & Repair & 69.08 & {\cellcolor[HTML]{95C095}} 5.01 & 56.82 & {\cellcolor[HTML]{E7C0C0}} -3.35 & 56.27 & {\cellcolor[HTML]{71AA71}} 8.92 & 25.50 & {\cellcolor[HTML]{EBC9C9}} -2.60 \\
 \midrule
\multirow{4}{=}{\parbox[t]{\hsize}{Qwen2.5-Coder-7B}} & Base Model & 52.92 & \cellcolor[HTML]{E6E6E6} {--} & 50.97 & \cellcolor[HTML]{E6E6E6} {--} & 17.83 & \cellcolor[HTML]{E6E6E6} {--} & 15.20 & \cellcolor[HTML]{E6E6E6} {--} \\
 & Generation & 58.22 & {\cellcolor[HTML]{71AA71}} 5.30 & 50.14 & {\cellcolor[HTML]{D9A0A0}} -0.83 & 15.88 & {\cellcolor[HTML]{F8E7E7}} -1.95  & 16.40 & {\cellcolor[HTML]{71AA71}} 1.20 \\
 & Completion & 19.77 & {\cellcolor[HTML]{D9A0A0}} -33.15 & 64.07 & {\cellcolor[HTML]{71AA71}} 13.10 & 1.39 & {\cellcolor[HTML]{D9A0A0}} -16.44 & 15.00 & {\cellcolor[HTML]{D9A0A0}} -0.3 \\
 & Repair & 55.71 & {\cellcolor[HTML]{AECFAE}} 2.79 & 51.25 & {\cellcolor[HTML]{F0F7F0}} 0.28 & 40.95 & {\cellcolor[HTML]{71AA71}} 23.12 & 15.00 & {\cellcolor[HTML]{D9A0A0}} -0.3 \\
\bottomrule
\end{tabularx}
\end{table*}

\textbf{Task-specific fine-tuning: strong within-task gains, weak transfer.}
Table~\ref{tab:model_comparison_finetune} reports our fine-tuning experiments.
In most cases, our fine-tuning reliably improves within-task model performance (\textit{e.g.}, repair-tuned models improve on repair), but we do not observe consistent improvements on other modalities, including LiveCodeBench.
Across all six settings, fine-tuning improves performance on the \emph{trained} modality (six improvements and no degradations).
For cross-task transfer, we observe a pattern consistent with no improvement in capabilities, similar to public checkpoints.
Across 18 cross-task comparisons (each fine-tuned checkpoint evaluated on three held-out benchmarks), we observe five improvements and 13 degradations.
The low scores for Qwen2.5-Coder-7B-Instruct fine-tuned on code completion reflect overfitting to the output format.

These results support our claim: reasonable fine-tuning with within-task evaluation (practical from an effort standpoint) may mislead practitioners into perceiving a "capability jump" that largely reflects task-specific specialization rather than transferable coding ability.
Notably, excluding output format failures, we observe no difference in cross-task transfer between our benchmarks and LiveCodeBench despite their widely different task distributions.
This cross-task regression matters for LLM and coding agent development reliability: a checkpoint optimized for one workflow may be worse on other ones.

\section{Discussion}
\subsection{The \textit{meaning gap} thesis}

Our experiments provide empirical support for the \emph{meaning gap}: performance on a popular coding benchmark can fail to predict performance on other coding tasks, even if they are closely related.
As a result, broad capability claims based on a narrow set of leaderboards are often overstated.

\textbf{Limited evidence of cross-benchmark rank reversals for foundation models.}
If a single benchmark is a good proxy for ``general coding capability,'' the relative ordering of models should be broadly stable across different coding evaluations.
In our limited comparison (Qwen2.5 vs.\ Qwen3), we observe a rank reversal: Qwen3-32B scores higher on SWE-bench, yet performs substantially worse than Qwen2.5-Coder-32B-Instruct models on our code completion benchmark.
This \emph{suggestive rather than definitive} evidence illustrates the core risk: leaderboard position on one benchmark can be an unreliable signal for another task that also corresponds to a facet of general coding capability.

\textbf{Post-trained checkpoints show near-universal failure of cross-task transfer.}
The meaning gap is more pronounced for post-trained checkpoints.
Across both community-released SWE-bench-optimized checkpoints and our fine-tuned variants, improvements are largely confined to the training/evaluation distribution, with little to no consistent gains on our Django benchmark suite and LiveCodeBench.
In several cases, performance severely degrades on other tasks (e.g., SWE-agent-LM-7B and OpenHands-LM-32B underperform their base models across multiple benchmarks, failing to follow instructions).
Moreover, failures are not uniform: some checkpoints follow instructions and formats reasonably well, yet still do not improve out-of-distribution, making the lack of transfer difficult to detect via qualitative ``sanity checks'' alone. 
As LiveCodeBench and our benchmark suite use very different data and test models on different tasks, we speculate this may mean lack of cross-task improvement on other coding tasks as well.

\textbf{Benchmarks capture coarse capability differences, but are brittle for fine-grained choices.}
We do not claim benchmarks are useless.
Across model \emph{tiers} (e.g., small vs.\ mid-size vs.\ frontier systems), stronger models tend to outperform weaker ones on most tasks, indicating that coding capabilities \emph{do} improve over time and scale.
The practical problem arises precisely where users most need guidance: choosing between models that are close in headline benchmark scores (\textit{e.g.}, selecting the "best 8B model") or, more importantly, checkpoints that are fine-tuned for a specific task.
In this regime, a model that edges ahead on SWE-bench can lag behind on completion or repair, so single-leaderboard rankings become unreliable guides for practitioners' and researchers' decisions.
This is where the meaning gap causes the most harm and where multi-task evaluation or task-specific human-in-the-loop studies are most necessary.

\subsection{Possible paths forward}

We see several complementary approaches to addressing the meaning gap, each with distinct tradeoffs.
We broadly divide them into four groups:
\textbf{(i)} single-score benchmarks,
\textbf{(ii)} diverse benchmark suites,
\textbf{(iii)} human-in-the-loop studies on real tasks, and
\textbf{(iv)} holistic open-ended evaluations.
Single-score benchmarks remain useful for fast iteration (regression tests, ablations), but are much weaker as final assessments or evidence for broad capability claims.

\textbf{Benchmark suites (beyond a single score).}
The most direct response to single-benchmark limitations is broader coverage: a suite spanning completion, editing, bug localization, question answering, and repository navigation.
MTEB~\cite{muennighoff2023mteb} exemplifies this for text embeddings, where diverse tasks constrain overgeneralization.
However, this approach faces two obstacles.

First, \textbf{benchmarks are high-effort community service} that is weakly rewarded, especially for maintenance, which \textbf{matters even more than creation}.
Each conference introduces many new benchmarks, but very few are ever widely adopted, and fewer are still maintained as models, data, and contamination risks evolve.
High-quality suites need sustained curation, robust evaluation harnesses, contamination monitoring, and regular refresh~\cite{bean2025measuring}.

We believe this is addressable \textit{institutionally}: targeted grants (including industry sponsorship), conference tracks and workshops that reward ongoing maintenance (\textit{e.g.}, new versions, repaired instances, and contamination audits), not just initial releases, treating benchmarks as first-class research artifacts.
Without recognizing benchmark maintenance as essential research infrastructure, the existence and support of benchmarks like SWE-Bench-Pro, SWE-rebench~\cite{badertdinov2025swe}, or LiveCodeBench depend on the chance and goodwill of a small number of individuals. 

The second obstacle is a \textbf{construct validity} problem: a benchmark suite is a good proxy for ``general coding capability'' only to the extent that it is built on a clear and correct taxonomy of coding skills and behaviours it aims to measure.
In addition to the general construct validity questions raised by~\cite{bean2025measuring}, a robust suite requires the community to answer (at least) the questions below:
\begin{itemize}
    \item \textbf{Define the construct:} what behaviours count as ``coding capability'' for the target use cases?
    \item \textbf{Choose domains:} what distributions matter (open-source vs.\ proprietary ecosystems~\cite{deng2025swe}), and how should we handle contamination risks across them?
    \item \textbf{Choose modalities and coverage:} which task types and programming languages (and language families) must be represented?
    \item \textbf{Decide aggregation:} how should we combine axes (tasks, repos, languages) to report an interpretable \emph{profile} rather than an over-compressed single score?
\end{itemize}
Developing such a taxonomy is substantial research, but other areas of science (\textit{e.g.}, psychometrics) suggest it is feasible: latent constructs are estimated through batteries of partially correlated tests, not single measurement.
We sketch how this taxonomy can look like in Appendix~\ref{app:taxonomy}. 
We stress this taxonomy is not ready to use, as creating usable taxonomy requires extensive empirical and qualitative work beyond the scope of our paper.

\textbf{Human-in-the-loop validation.}
A complementary approach is evaluation on genuinely real-world tasks with human judgment.
For SE tasks, this could mean sampling open GitHub issues or internal tickets and collecting structured assessments from maintainers or domain experts.
This can be uncontaminated by design (new issues are not in training data) and better reflect practical distributions, including cases that are underspecified, unsolvable with available context, or unnecessary to address.
Moreover, human in the loop validation can also assess the perception of a coding agent, which affects the efficiency of human-agent collaboration.
However, human evaluation has its own failure modes: expense, limited scalability, inconsistent standards across projects, and the risk of preferring style over substance~\cite{wu2025style}.

\textbf{Holistic open-ended evaluation.}
A more radical departure from fixed tasks is holistic evaluation in open-ended scenarios.
E.g. to succeed in Anthropic's Project Vend~\cite{projectvend} a model has to integrate many skills under events that no benchmark designer would anticipate.
For coding, analogous setups can be controlled hackathons or bug-squashing days where teams are randomly assigned AI assistants, with models judged on outcomes \emph{and} user experience.

These can be run as model--vs--model competitions with standardized rules, potentially sponsored by model developers. 
The results can be reported as a public leaderboard if the protocol is transparent and robust enough to avoid becoming another easily optimized proxy.
While costly, a holistic evaluation targets the meaning gap directly: it measures performance when tasks are underspecified, requirements drift, and success requires adaptive problem-solving.

\textbf{Evaluation approaches in practice.}
These approaches are complementary.
For frontier foundation models, holistic assessments (and carefully designed human studies) should be the most informative.
For research on incremental techniques and smaller models, benchmark \emph{suites} should provide a practical and scalable way to detect narrow vs. broad improvements. 
Single-score benchmarks are useful for rapid iteration, but should not be treated as decisive evidence.
For narrow practical tasks such as fine-tuning a model on a private repository for internal usage, task-specific human-in-the-loop setups should be the most actionable:  general leaderboards are unreliable guides for models close in capability. 
This approach can surface narrow failure modes that public benchmarks miss.

\section{Alternative views}
In this paper, we propose a three-pronged evaluation approach: holistic assessment of frontier models, multi-task benchmarks for academic research, and task-specific user studies for narrow practical tasks. 
However, there are alternative views on model evaluation, which we present and discuss below.

Jordan et al.~\cite{jordan2024position} argue that rigorous benchmarking can be prohibitively expensive and propose using \textit{scientific testing} instead to understand how algorithms work.
While we agree that mechanistic understanding is valuable in academic research, we believe benchmarking model performance outside controlled experiments is equally necessary for algorithm evaluation—complementing rather than replacing scientific testing.
To paraphrase a famous statement, ``all DL approaches are not general, but some are useful'', and benchmarking is how one can estimate the limits of usefulness.

A related alternative is qualitative evaluation, exemplified by Yang et al.~\cite{yang2025swe}, which examines how and why models fail, what distinguishes success from failure, and what patterns emerge across cases.
Such analysis reveals capability boundaries invisible in summary statistics—whether models fail on long-range dependencies, specification understanding, or semantic correctness despite syntactic validity.
We view qualitative analysis as a valuable complement to quantitative evaluation, not a replacement, and believe qualitative findings can inform better benchmark design that addresses common failure modes.
However, the need for reliable quantitative comparison remains.

Another possible counterargument is that the evaluation system is self-correcting. 
For example, the emergence of SWE-bench and LiveCodeBench addresses the limitations of HumanEval: limited task scope and contamination, respectively. 
We are less optimistic: benchmark gains are now read as capability gains, masking failures to transfer and creating selection pressure for Goodhart's law-like effects to appear~\cite{manheim2018categorizing}.
It also takes time for the community to change the benchmark of choice: while first concerns about leaks and test quality in SWE-bench first appeared in 2024~\cite{aleithan2024swe}, SWE-bench is still the most popular benchmark for evaluating LLMs on SE tasks.
One of the goals of our contribution is to show the limitations of self-correction feedback.

\section{Conclusions: what should be done}
In this position paper, we argue that there is no ``one size fits all'' approach to DL-for-code model evaluation, and that using a single or a limited number of benchmarks to assess model performance does not provide sufficient information about model capabilities. 
We suggest using four complementary approaches: \textit{single-score benchmarks} for iteration, \textit{benchmark suites} for capability profiles, \textit{human-in-the-loop studies} for narrow deployments, and \textit{holistic open-ended evaluations} for frontier models.
To make benchmark suites reliable, we suggest treating them as research infrastructure: funding creation and long-term maintenance (e.g., targeted grants and industry sponsorship) and rewarding versioned updates via dedicated conference tracks. 
We also urge the community to ground benchmark suites in a shared capability taxonomy that covers tasks, domains, languages, and aggregation to ensure construct validity, and acknowledge the limitations of the benchmarking methods for the task at hand.
Finally, we also call on DL for code researchers working on training recipes and coding agents to explicitly report what capabilities transfer across tasks and to characterize measurement limitations of their evaluations. 

\bibliography{iclr2026_conference}
\bibliographystyle{icml2026}

\appendix
\section{Coding capability taxonomy illustration}
\label{app:taxonomy}
To illustrate what a coding task taxonomy could look like, we include a sample taxonomy, see Tables~\ref{tab:software-engineering-task-families-part1}, ~\ref{tab:software-engineering-task-families-part2}.
This taxonomy was created in a discussion between two of the paper authors and is \textbf{not} grounded in qualitative and quantitative research. 
As such, it is merely an illustration of what kind of a taxonomy would be helpful to benchmark creators so that they can ensure the construct validity of their benchmarks and assess their limitations.
We urge \textbf{not} to use this taxonomy as is and call for research community to develop a proper taxonomy with research methods from other areas of science such as psychometrics and social anthropology.

\begin{table*}[t]
\centering
\small
\begin{tabular}{p{0.18\textwidth} p{0.34\textwidth} p{0.42\textwidth}}
\toprule
\textbf{Task family} & \textbf{Individual tasks} & \textbf{How to gather} \\
\midrule

\multicolumn{3}{l}{\textbf{Design \& planning}} \\
\midrule
Architecture \& planning
&
API, schema, module design, subsystem decomposition
&
Mine ADRs (Architecture Decision Record), RFCs (Request for Comments), epic issues, and linked PR series; author tasks from real design docs
\\

Requirement analysis
&
Problem decomposition, requirement verification 
(e.g., checking for non-contradiction)
&
Mine development roadmaps and issue trackers; mine design discussions for requirements verification
\\

\midrule
\multicolumn{3}{l}{\textbf{Understanding \& communication}} \\
\midrule
Repo understanding
&
Retrieve relevant files; answer a repo question
&
PRs, issues, commit-message to changed-files pair, comments or discussions in GitHub, synthetic generation
\\

Documentation \& communication
&
PR summary, commit message, code explanation, 
README.md, changelogs, project documentation
&
Mine accepted PR descriptions, docs diffs, maintainer answers, README updates, project documentation and changelogs
\\

Data wrangling
&
Input and output prediction for single and many-step pipeline
&
Create diverse outputs via fuzzing testing on various data pipelines
\\

\midrule
\multicolumn{3}{l}{\textbf{Coding \& feature development}} \\
\midrule
Repo-level implementation / features
&
Add endpoint/page/workflow across multiple files
&
Mine enhancement/feature PRs; filter out bug fixes
\\

Repo-level synthesis \& completion
&
Fill a function/class/file-sized hole using info on the repo
&
Mask historical commits; create repo hole-filling tasks with hidden tests. Hide method implementations, synthetically generating docs for it.
\\

Local synthesis \& completion
&
Fill a function/class/file-sized hole
&
Use fresh contest problems
\\

\midrule
\multicolumn{3}{l}{\textbf{Validation, diagnosis \& feedback}} \\
\midrule
Debugging \& repair
&
Reproduce bug, localize root cause, patch it
&
Mine bug reports with fix commits, CI failures with later fixes, stack traces, and failing tests
\\

Testing \& verification
&
Generate unit, integration, regression tests
&
Hide existing tests, mine test-adding commits, or preserve new regression tests from bug-fix PRs
\\

Review \& critique
&
Review a diff and identify concrete issues, decide whether to accept PR
&
Mine merged PRs with human review comments and suggestions
\\

\bottomrule
\end{tabular}
\caption{Task families, individual software-engineering tasks, and possible data-gathering strategies, part 1.}
\label{tab:software-engineering-task-families-part1}
\end{table*}

\begin{table*}[t]
\centering
\small
\begin{tabular}{p{0.18\textwidth} p{0.34\textwidth} p{0.42\textwidth}}
\toprule
\textbf{Task family} & \textbf{Individual tasks} & \textbf{How to gather} \\
\midrule

\multicolumn{3}{l}{\textbf{Codebase evolution \& adaptation}} \\
\midrule
Refactoring
&
Behavior-preserving extract, rename, move, modularize
&
Mine refactor-only commits/PRs using detectors plus post-filtering
\\

Migration \& compatibility
&
Upgrade runtime, library, framework, API, language version
&
Mine upgrade PRs, version bumps, deprecation-fix commits, lockfile changes
\\

Code translation
&
Switch codebase to another programming language
&
Mine code transformation PRs
\\

\midrule
\multicolumn{3}{l}{\textbf{Hardening \& optimization}} \\
\midrule
Security hardening
&
Fix a vulnerability or produce secure repo-level code
&
Mine CVEs (Common Vulnerabilities and Exposures), advisories (vulnerability reports), Static/Dynamic Application Security Testing findings, and security-fix PRs
\\

Performance optimization
&
Speed up code while preserving behavior on repo-level tasks
&
Mine performance-improving PRs or pair tasks with optimized baselines. Synthetically generate the code optimization.
\\

Local performance optimization
&
Speed up code while preserving behavior on stand-alone tasks
&
Use fresh contest problems, mine solutions with various time/memory consumption
\\

\midrule
\multicolumn{3}{l}{\textbf{Delivery infrastructure \& lifecycle}} \\
\midrule
Environment / CI / maintenance
&
Set up repo, fix failing CI, keep code green over iterations
&
Mine setup docs/workflows, failing CI snapshots, and long commit histories
\\

\midrule
\multicolumn{3}{l}{\textbf{Tool and skill usage}} \\
\midrule
Codebase navigation
&
Parsing multi-file dependencies, finding relevant method
&
Create synthetic datasets with code parsers such as tree-sitter
\\

Tool usage
&
Running correct commands in terminal, calling external API, using tools such as Git
&
Mine repos with API usages, mine tool and terminal usage patterns from manuals and educational tasks
\\

Style following
&
Following code style, following documentation guides
&
Mine documentation guides, code style guides such as PEP, diffs before / after linter usage
\\

\bottomrule
\end{tabular}
\caption{Task families, individual software-engineering tasks, and possible data-gathering strategies, part 2.}
\label{tab:software-engineering-task-families-part2}
\end{table*}
\section{Django Benchmark Suite Details}
\label{app:benchmark}

In this section we provide a detailed description of our benchmark suite construction, expanding on the methodology outlined in Section~\ref{sec:experiments}. 

\subsection{Repository Selection and Leakage Control}
\label{app:repo}
We base our benchmark suite on a single, frozen snapshot of Django at version 4.0.4, commit hash $89807fbde8b7b17d00434bc4695535855e96fe77$, dated 11th Apr 2022.
We select this snapshot for several reasons.
We choose Django because it constitutes 46\% of SWE-bench Verified, allowing us to test a stronger version of the meaning gap thesis: if external checkpoints perform well on SWE-bench Verified, they should perform well on Django issue resolution by extension. 
Therefore, failure to improve on our benchmark suite indicates failure of cross-task transfer, suggesting even stronger limitations on performance in truly generalized setups with non-SWE-bench repositories and tasks other than issue-resolution.
We choose Django 4 specifically because it represents the largest share of major Django releases in SWE-bench Verified, and Django introduces breaking changes between major versions. 
Finally, we select this particular commit because it corresponds to the last release before Django 4.1 development began.
We hypothesize this version is most likely to have addressed issues from the initial Django 4.0 release while remaining unaffected by parallel development of the next version.
As we have no prior knowledge of Django development timeline, this motivation allows us to choose snapshot that is likely stable and representative.

\subsection{Train-Test Split}
\label{app:split}
After cloning the repository, we split the source files into train and test splits while tracking inheritance structure.
For each class, we place the class, its ancestors, and its descendants into the same split.
This prevents cross-split contamination via inherited methods, mixins, and abstract base classes.
The code outside classes, such as standalone functions or module-level imports, is assigned to a particular file and never duplicates across splits.
We do not use tests (code that lies in the \texttt{tests} folder) for training or inference purposes.
Table~\ref{tab:dataset_stats} summarizes the key statistics of our training and testing datasets.

\begin{table}[h]
\centering
\caption{Dataset statistics for training and testing splits.}
\label{tab:dataset_stats}
\begin{tabular}{lcc}
\toprule
\textbf{Statistic} & \textbf{Train} & \textbf{Test} \\
\midrule
Number of methods & 3,180 & 359 \\
Avg.\ method length (lines) & 10.27 & 9.25 \\
Median method length (lines) & 4 & 4 \\
Number of unique files & 425 & 58 \\
Number of inheritance chains & 1,585 & 184 \\
\midrule
\makecell[l]{Avg.\ generated documentation \\ length (chars)} & 243.87 & 236.60 \\
\makecell[l]{Median generated documentation \\ length (chars)} & 233.00 & 229.00 \\
\makecell[l]{Methods w/o original doc.} & 74.28\% & 78.27\% \\
\bottomrule
\end{tabular}
\end{table}

The training dataset is approximately 9$\times$ larger than the testing dataset, with similar average method lengths across both splits, indicating consistent complexity distribution.
The number of inheritance chains reflects our splitting strategy that keeps related classes together to prevent cross-split contamination.

\subsection{Documentation Synthesis}
\label{app:docs}
Django project documentation is uneven: some methods are documented thoroughly, while others have minimal or no documentation at all.
To create a uniform evaluation setup, we synthesize standardized docstrings for every method of every class using DeepSeek-R1.
We enforce a rigid schema for all generated docstrings:
\begin{itemize}
    \item \textbf{What it does:} One or two sentences describing the method's purpose and behavior.
    \item \textbf{Main inputs:} A one-sentence description of the method's parameters and their expected types or constraints.
    \item \textbf{Outputs and side effects:} A one-sentence description of the return value, plus description of expected failure modes (such as raising an exception in certain scenarios).
\end{itemize}
We keep the original comments present in the source code and merge them with generated docstrings whenever applicable, preserving any domain-specific information the original developers included.

\subsection{Test Suite and Method-Test Alignment}
\label{app:eval}
We build a Django test environment that runs the official 4.0.4 test suite.
For each method in our benchmark, we identify the tests that determine its correctness through a systematic process:
\begin{enumerate}
    \item For each target method, we replace its body with a stub implementation \texttt{return None}.
    \item We run all Django tests and record which tests fail due to the stubbed method.
    \item We map the failed tests to the method in question.
    \item At evaluation time, we run only these mapped tests to check method validity.
\end{enumerate}
Methods for which no tests failed when stubbed with \texttt{return None} are excluded from the benchmark, as their correctness cannot be verified through testing.

For a generated method to be considered correct, it must pass all mapped tests.
This approach ensures that we evaluate only the functionality directly related to each target method, rather than running the entire test suite for every evaluation.

\section{Additional evidence for SWE-bench limitations}
\subsection{Error-flow analysis}
\label{app:error-flow}
To support our position on possible lack of cross-task transfer, we performed a per-example error-type flow analysis that traces how each test outcome changes when a base model is replaced by its fine-tuned counterpart.

\begin{figure}[h]
    \centering
    \includegraphics[width=\linewidth]{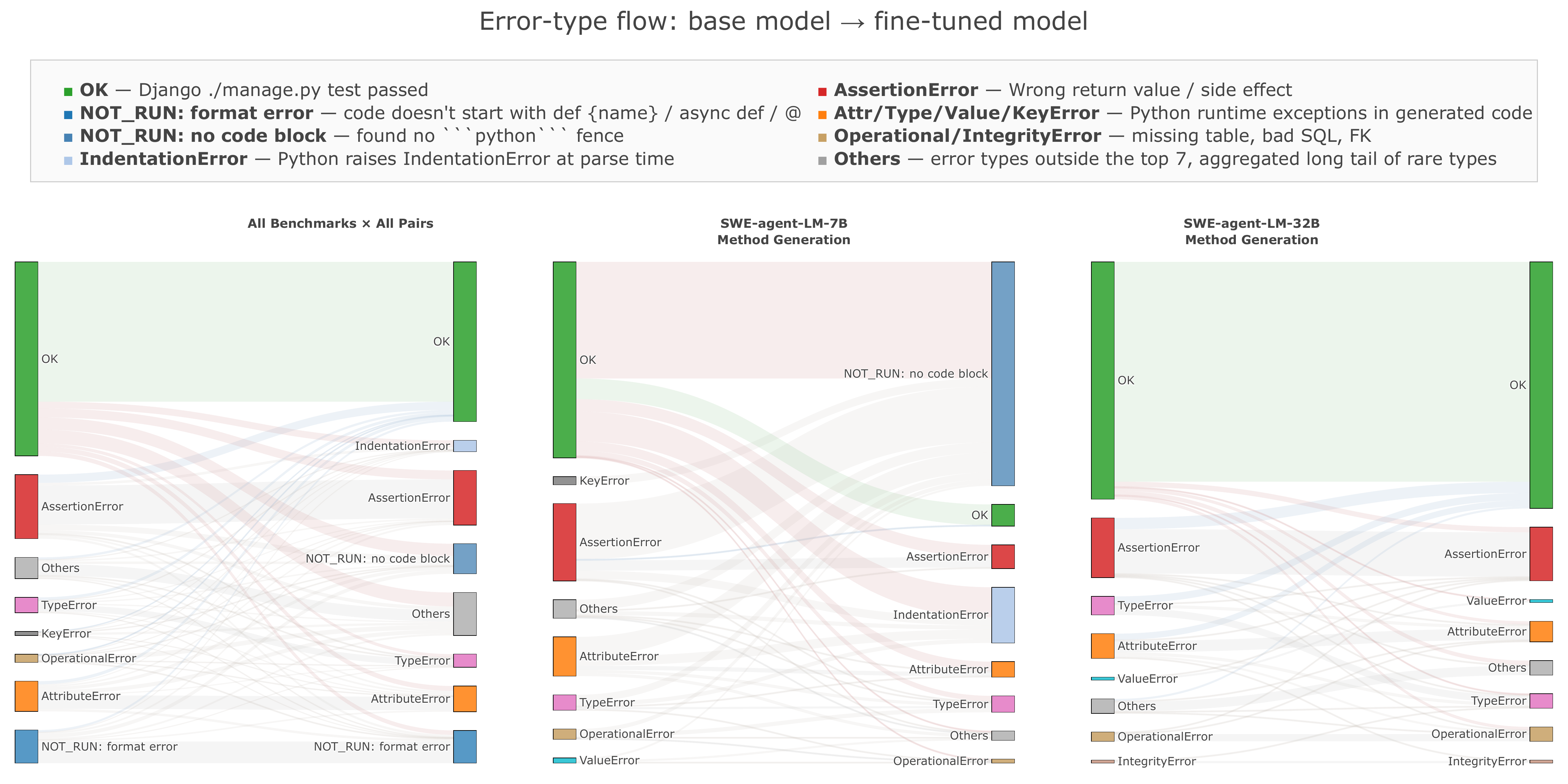}
    \caption{Per-example error type flow Sankey diagram}
    \label{fig:sankey}
\end{figure}

Across all seven model pairs and three benchmarks (7 539 examples), fine-tuning drops the OK rate from 52.4\% to 43.1\%. 
The modal dominant regression path is OK → NOT\_RUN: fine-tuned models emit agent-style output (XML tags, markdown fences, preamble code) that the harness cannot parse. 
In total, 47.2\% of regressions come from format or parsing error (NOT\_RUN + IndentationError), and 52.8\% of regressions are caused by essential mistakes such as AssertionError or SyntaxError. 
Thus, the impact of coding capability degradation and failure to follow the format are compatible. 
Model capacity determines whether this overfitting occurs. Comparing SWE-agent-LM-7B and -32B (both fine-tuned on SWE-smith and have $>13\%$ improvements on SWE-bench) on Method Generation:
\begin{itemize}
    \item 7B (52.9\% → 5.8\% OK): the largest flow is OK → NOT\_RUN (113 examples). The model wraps output in [object Object]…[object Object] tags instead of a bare def block; 54 further examples acquire IndentationError.
    \item 32B (64.1\% → 66.6\% OK): the dominant flow is OK → OK (213); 11 previously failing cases are now fixed. The larger model compartmentalises agent formatting and retains the output format expected by the harness. On the per-model panels it is a long tail of rare types (each two examples or less, e.g. NameError, ImportError, RecursionError).
\end{itemize}
To sum up, while checkpoints fine-tuned on agentic trajectories often fail due to format non-adherence, failure modes are numerous and their distribution is long-tailed. 
More than half of regressions are due to conceptual errors in model generations. 
These failure modes cannot be observed if a checkpoint is evaluated on a single benchmark such as SWE-Bench Verified, supporting our position that current evaluation practices may obscure problems with cross-task transfer.

\subsection{Impact of agentic scaffolding}
\label{app:scaffolding}
To check the impact of agentic scaffolding, we compare the SWE-agent checkpoint to raw models in an agentic scaffolding setup on method generation task, see Table~\ref{tab:swe-agent-scaffolding}. 
This setup is organized as follows.
SWE-agent is placed into Django 4.0.4 repository. 
It gets a prompt with the file path to the target file and problem statement (description of target method to generate and system prompt to generate it). 
The agent does not get the file contents directly. 
The target file is missing the method that the agent should implement, similar to no-scaffolding setup. The agent gets ten steps to implement it. 
The agent can only submit solution once, and the solution is either submitted by a special tool call, or submitted automatically when the step cutoff is hit.

\begin{table}[t]
\centering
\small
\begin{tabularx}{\columnwidth}{Xcc}
\toprule
\textbf{Model / scaffolding} & \textbf{Scaffolding, 10 steps} & \textbf{No scaffolding} \\
\midrule
Qwen2.5-Coder-7B-Instruct & 0.0 & 52.92 \\
SWE-agent-LM-7B & 24.51 & 5.85 \\
Qwen2.5-Coder-32B-Instruct & 8.59 & 64.07 \\
SWE-agent-LM-32B & 39.62 & 66.57 \\
\bottomrule
\end{tabularx}
\caption{Model performance with and without SWE-agent scaffolding.}
\label{tab:swe-agent-scaffolding}
\end{table}

When both SWE-agent checkpoints and the raw model are put into the same scaffolding, the fine-tuned checkpoints outperform the raw model. 
However, both for raw models and SWE-agent-LM-32B checkpoint removal of scaffolding further improves the results in ten-step setup.
To check further if ten steps are insufficient and whether the agent runs out of steps, we evaluate SWE-agent-LM-32B checkpoint in agentic scaffolding with 75-step limit. 
We consider this particular checkpoint as it has the smallest difference with the raw model non-agentic performance, and we do not run other checkpoints due to budget limitations. 
In this setup, SWE-agent-LM-32B scores $63.51$, so it takes 75 steps in an agentic scaffolding for a particular SWE-agent checkpoint to match a non-agentic setup for either SWE-agent checkpoint or raw model.
The difference between ten and 75-step scaffolding performance highlights that agentic setup requires multiple tries to solve the task and suggests that in ten-step setup running out of steps is the key failure mode. 

All in all, this indicates that regardless of the fine-tuning approach some tasks are better solved without an agentic scaffolding.
The failure of SWE-agent-LM-32B checkpoint to outperform the raw model in non-agentic setup further supports our point on the lack of cross-task transfer.
We stress that creators of SWE-agent checkpoints could not have checked this without creating additional benchmarks (a significant research and technical effort), so this reflects a failure of the evaluation ecosystem, not of the researchers themselves. 

\section{Task Descriptions}
\label{app:tasks}

We create three distinct benchmarks from our Django repository snapshot.
Each benchmark is accompanied by a corresponding fine-tuning dataset for instruction tuning on the same task format.
Below we describe the input-output format and evaluation protocol for each task.

\subsection{Method Generation}

In the method generation task, the model is given a method specification and must generate the complete method body.

\paragraph{Input format.}
The input consists of the following components in order:
\begin{enumerate}
    \item A task description instructing the model to generate the missing method body.
    \item The target class name and the declaration (signature) of the method to be written.
    \item The synthesized docstring for the target method, following our standardized schema.
    \item The entire source file content, with the target method body removed. We reorganize non-target code blocks to place the missing method at the end of the file, but otherwise preserve the original file structure.
\end{enumerate}

\paragraph{Output format.}
The expected output is the full method code (including its signature) wrapped in a \texttt{```python} code block.
We extract the code from the last code block in the model's response and ignore any additional text or code blocks.
Appending the correct output to the provided file context should yield a syntactically valid and functionally correct file.

\paragraph{Evaluation.}
We evaluate with the pass@1 metric at temperature $T=0$ (greedy decoding).
For each generated method, we insert it into the source file and run only the mapped Django tests for that method.
A solution is considered correct if and only if all mapped tests pass.

\subsection{Method Completion}

The method completion task is similar to method generation, but provides partial implementation context instead of a docstring.

\paragraph{Input format.}
The input consists of:
\begin{enumerate}
    \item A task description instructing the model to complete the method.
    \item The source file content up to and including approximately half of the target method body.
    \item The model must generate the remaining portion of the method.
\end{enumerate}
Unlike method generation, we do not provide the synthesized docstring. 
Instead, the model must infer the method's intended behavior from the partial implementation and surrounding code context.

\paragraph{Output format.}
The model should generate only the completion---the remaining lines of the method body that, when concatenated with the provided prefix, form a complete and correct method.
We extract the generated code and append it to the provided partial method.

\paragraph{Evaluation.}
We use the same evaluation protocol as method generation: the completed method is inserted into the source file and validated against the mapped Django tests using pass@1.

\subsection{Program Repair}

In the program repair task, the model receives a file with a broken method and must identify and fix the bug.
To create plausible buggy methods, we use a cascading approach: we first collect failing outputs from three runs of Qwen3-32B (89 samples), then from Qwen3-8B for tasks solved by the larger model (83 samples), and finally from Qwen3-0.6B (126 samples). For the remaining 61 tasks solved by all models, we shuffle non-control tokens within methods.

The idea of creating synthetic program repair benchmark follows~\cite{ouyang2024benchmarking}, our approach for creating LLM-generated corrupted methods is analogous to ``LM-rewrite'' approach for SWE-smith dataset creation~\cite{yang2025swe}, and our fallback approach of token shuffling and method candidate shuffling is inspired by the ``Procedural modification'' approach of SWE-smith.

\paragraph{Input format.}
The input consists of:
\begin{enumerate}
    \item A task description explaining that one method in the file is broken and needs to be fixed.
    \item The complete source file containing one method with an introduced bug.
    \item The full stack trace from the Django test suite when run against the broken method, indicating which tests failed and where the error occurred.
\end{enumerate}
The model must analyze the stack trace, identify the broken method, understand the nature of the bug, and generate a corrected version.

\paragraph{Output format.}
The model should return only the fixed method implementation. We extract it and substitute the broken method in the source file.

\paragraph{Evaluation.}
We run the mapped tests for the originally broken method.
A repair is considered successful if all mapped tests pass.
We evaluate with pass@1 at temperature $T=0$.

\subsection{Benchmark Verification with Proprietary Models}
\label{app:benchmark-verification}

We evaluate a diverse set of state-of-the-art proprietary LLMs on our benchmark suite.
Our goal is not to rank these foundation models, but to verify our benchmarks are well-posed and solvable by the best LLMs.

We evaluate all the models under a setup identical to the other evaluations we did. 
The only exception is the temperature for the GPT-5 model, as we could not set $T=0$ for it.
We use evaluation protocol identical to the one we use for other model checkpoints, and observe no issues with context sizes, rate limits or any other problems that can arise when using proprietary models through API calls.

\begin{table}[t]
\centering
\begin{tabular}{|l|l|l|l|l|}
    \hline
        Model & CG & CC & PR \\ \hline
        OpenAI GPT-4.1 & 76.60 & 64.90 & 61.84 \\ \hline
        OpenAI GPT-5 & 84.68 & 77.99 & 79.67 \\ \hline
        Google Gemini 2.5 Pro & 83.01 & 51.25 & 67.97 \\ \hline
        Google Gemini 3 Pro & 86.35 & 78.27 & 61.00 \\ \hline
        Google Gemini 3 Flash & 85.79 & 73.26 & 62.95 \\ \hline
        Anthropic Claude Haiku 4.5 & 74.65 & 58.77 & 61.56 \\ \hline
        Anthropic Claude Sonnet 4.5 & 82.73 & 77.44 & 69.36 \\ \hline
        Anthropic Claude Opus 4.5 & 88.02 & 76.60 & 78.27 \\ \hline
    \end{tabular}
    \caption{Performance of Proprietary LLMs}
    \label{tab:proprietary-results}
\end{table}

\paragraph{Result Analysis.}
The results in Table~\ref{tab:proprietary-results} provide several insights into the properties of our benchmarks.

First, all three tasks are clearly solvable by contemporary high-capability models, with top-performing systems exceeding 80\% pass@1 on method generation and method completion, and almost reaching this score for program repair. 
This indicates that the benchmarks are well-posed and solvable.

Second, the results align with expected capability orderings within model families.
GPT-5 outperforms GPT-4.1 across all tasks. 
Claude Opus 4.5 outperforms Sonnet 4.5 on two tasks out of three, and Sonnet  outperforms Haiku 4.5 on all three tasks. 
Gemini 3 Pro and Flash outperform Gemini 2.5 Pro on code generation and code completion. These orderings are consistent with the models' relative positions on other established benchmarks.

Third, we observe a consistent ordering of task difficulty across models. 
Method generation (CG) is generally easier than method completion (CC) and program repair (PR), similarly to what we observe for the open-weights models and post-trained checkpoints. 

Fourth, the table reveals that even frontier models exhibit uneven capability profiles across tasks, with various models having different strengths and weaknesses.
For example, GPT-5 performs significantly better than Gemini 3 Pro on code completion, but is not better on code generation and program repair.

All these observations support our claim that the proposed benchmarks are well-posed, pass the sanity checks, and test distinct aspects of LLM coding capabilities, which are, although connected, still distinguishable.
Thus, we can use them as a diagnostic tool to support our hypothesis on meaning gap and risk of cross-task transfer.

\section{Fine-tuning Setup}
\label{app:finetuning}
We fine-tune Qwen2.5-Coder-7B-Instruct and Qwen2.5-Coder-32B-Instruct on each of the three task modalities described above.
We choose these models because most of the external checkpoints we evaluate were built on top of one of these two base models, enabling direct comparison.

\textbf{We do not claim that our fine-tuning experiments extract all possible value from the repository data.}
Our goal is to demonstrate how a checkpoint created with a reasonable, standard fine-tuning approach performs across different tasks, and not to achieve state-of-the-art results through extensive hyperparameter optimization.
Our configuration represents a practical setup that practitioners might use when adapting models to new domains, revealing how task-specific training affects cross-task generalization.

\subsection{LoRA Configuration}

We use Low-Rank Adaptation (LoRA) for parameter-efficient fine-tuning with the following configuration:
\begin{itemize}
    \item \textbf{Rank:} 128
    \item \textbf{Alpha:} 128
    \item \textbf{Dropout:} 0.05
    \item \textbf{Target modules:} All attention and MLP projection layers (\texttt{q\_proj}, \texttt{k\_proj}, \texttt{v\_proj}, \texttt{o\_proj}, \texttt{gate\_proj}, \texttt{up\_proj}, \texttt{down\_proj})
\end{itemize}
We target all linear layers in the transformer architecture to maximize the model's capacity to adapt to the new task distribution while keeping the number of trainable parameters manageable.

\subsection{Training Hyperparameters}

\begin{itemize}
    \item \textbf{Optimizer:} AdamW with weight decay of 0.01
    \item \textbf{Learning rate:} $1 \times 10^{-5}$
    \item \textbf{Learning rate scheduler:} Cosine annealing
    \item \textbf{Warmup steps:} 30
    \item \textbf{Maximum gradient norm:} 1.0
    \item \textbf{Number of epochs:}  3
    \item \textbf{Batch size:} 
        \begin{itemize}
            \item 1 per device with gradient accumulation over 8 steps (effective batch size of 8) for Qwen2.5-Coder-7B-Instruct
            \item 4 per device with gradient accumulation over 8 steps (effective batch size of 32) for Qwen2.5-Coder-32B-Instruct
        \end{itemize}
\end{itemize}

In every case, we evaluate the last epoch checkpoint.

These settings are \emph{not} the result of extensive hyperparameter search.
We deliberately use standard, reasonable defaults to demonstrate that even without careful tuning, single-task fine-tuning produces models with improved in-distribution performance but limited cross-task transfer.
This supports our main thesis about the meaning gap in benchmark evaluation and risks of missing it in single-benchmark evaluation setup.

\section{Prompt Templates}
\label{app:prompts}

This section provides the exact prompt templates used for each task.
Each task uses a system prompt that establishes the assistant's role and output constraints, followed by a user prompt template that provides the specific task context.
Placeholders in curly braces (e.g., \texttt{\{file\_content\}}) are replaced with actual values at inference time.

\onecolumn
\lstdefinestyle{prompt}{
  basicstyle=\ttfamily\footnotesize,
  breaklines=true,
  breakatwhitespace=false,
  frame=single,
  framesep=2pt,
  xleftmargin=0pt,
  xrightmargin=0pt,
  aboveskip=5pt,
  belowskip=5pt,
  columns=fullflexible,
  keepspaces=true,
  breakindent=0pt,
  postbreak=\mbox{\textcolor{gray}{$\hookrightarrow$}\space},
}

\subsection{Method Generation Prompts}
\label{app:prompts:method_gen}

\paragraph{System prompt.}\mbox{}\\[-0.5\baselineskip]
\begin{lstlisting}[style=prompt]
You are a helpful Python code-generation assistant.

**Output rules**

* Return **only** the target method (signature + body) inside **one** fenced code block labeled `python`.
* Do **not** include the containing class or any other code.
* Do **not** add imports or top-level definitions.
* No prose or explanations outside the code block. Comments are allowed **inside** the code.

**Signature & decorator rules**

* Start with the **exact** method declaration provided in `{method_declaration}` (name, parameters, defaults, `async`, return annotation) -- do **not** change it.
* You may add decorators **immediately above** the declaration **only if necessary** for correctness **within a class**:

**Implementation rules**

* Match the project's style (typing, error handling, logging, docstrings).
* Use only modules already imported in the file; if a tiny helper is required, define it **as a nested function inside the method**.
* Access only attributes that exist in the class per the provided file unless the description explicitly introduces them.
* Prefer clear, test-ready logic. Avoid placeholders like `...`, `pass`, or `raise NotImplementedError` unless explicitly required.
* Ensure the method is self-contained and syntactically valid.

**Format**

```python
# method declaration (with an allowed decorator if necessary) followed by the method body only
```
\end{lstlisting}

\paragraph{User prompt template.}\mbox{}\\[-0.5\baselineskip]
\begin{lstlisting}[style=prompt]
Generate the implementation of the method `{method_name}` for class `{class_name}` according to the description.
**Return only this method** -- start with the **exact** declaration shown below (do not change it). If a decorator is necessary for correctness inside the class, place it on the line above the declaration. Put everything inside a single `python` code block.
Method declaration:

```python
{method_declaration}
```

Method description:
{method_description}

File context:

```python
{file_content}
```
\end{lstlisting}


\subsection{Method Completion Prompts}
\label{app:prompts:completion}

\paragraph{System prompt.}\mbox{}\\[-0.5\baselineskip]
\begin{lstlisting}[style=prompt]
You are a helpful Python code-completion assistant.

**Output rules**

* Return **only** the new code (the continuation), inside **one** fenced code block labeled `python`.

* Do **not** repeat any part of the provided file.

* Do **not** include explanations or text outside the code block.

* The output will be directly inserted into the end of the file, so the code must be correctly formatted, and must **preserve existing indentation levels** to avoid issues. The file from the context, plus the generated code should combine into a correctly working Python file

* Comments are allowed **inside** the code, sparingly.

**Completion rules**

* Continue from the end of the provided snippet, completing the **currently open function/method** until the file is valid and the method is fully implemented.

* Do **not** generate method declaration or existing body, only the continuation.

* Do **not** modify prior lines or signatures; do **not** introduce new top-level code.

* Match existing style (typing, naming, docstrings, error handling, logging).

* Use only already-imported modules and builtins. If a helper is required, implement it **inside** the completed method (local/nested).

* Ensure the result is syntactically valid, self-contained within the completed region, and aims to pass tests.

* Avoid placeholders like `...`, `pass`, `raise NotImplementedError` unless they are the intended final behavior.

**Format**

```python
# your continuation only
```
\end{lstlisting}

\paragraph{User prompt template.}\mbox{}\\[-0.5\baselineskip]
\begin{lstlisting}[style=prompt]
You are given a Python file with an **incomplete method**. Complete **only** the missing continuation so that the file becomes valid and the method is fully implemented.

**Return only the new code**, inside one fenced `python` code block. **Do not** repeat any existing lines from the file. Do **not** include explanations or examples.

File content:

```python
{file_content}
```
\end{lstlisting}

\subsection{Program Repair Prompts (With Test Output)}
\label{app:prompts:repair_test}

\paragraph{System prompt.}\mbox{}\\[-0.5\baselineskip]
\begin{lstlisting}[style=prompt]
You are a helpful Python program-repair assistant.

**Your task**

* From the provided file and test output, identify the **single** incorrect method and produce a fixed implementation.

**Output rules**

* Return **only** the repaired method (exact signature + body) inside **one** fenced code block labeled `python`.

* Repair exactly one method in the provided file that is most likely incorrect.

* Do **not** include the class wrapper, imports or any other code.

* Do **not** include explanations, prose, or examples outside the code block.

* Comments are allowed **inside** the code and after signature.

**Repair rules**

* Keep the method's **signature exactly as in the file** (name, parameters, defaults, `async`, return annotation). Do **not** change it.

* Do **not** add or change decorators.

* Do **not** introduce new imports or top-level definitions.

* Use only symbols already available in scope; if a tiny helper is needed, define it **as a nested function inside the method**.

* Follow the project's style (typing, error handling, logging).

* Use the **test output** to guide the repair; ensure the fix addresses failing cases.

* Avoid placeholders like `...`, `pass`, or `raise NotImplementedError` unless that is the intended behavior in the test outputs.

* If many methods look suspicious, fix the one that best explains the failing tests.

**Output format**

```python
# exact method declaration from the file, followed by the repaired body
```
\end{lstlisting}
\pagebreak
\paragraph{User prompt template.}\mbox{}\\[-0.5\baselineskip]
\begin{lstlisting}[style=prompt]
You are given a complete Python file. Exactly **one** method is incorrect. You also receive the failing test feedback.

Identify the incorrect method and output its **repaired implementation**.

**Return only this method**, starting with the **exact signature as it appears in the file** (no class wrapper, no comments), inside a single `python` code block.

Test output (failing tests):

```
{test_output}
```

Full file context:

```python
{file_content}
```
\end{lstlisting}

\subsection{Examples of Synthesized Documentation}
\label{app:prompts:doc_examples}

This section provides examples of documentation synthesized by DeepSeek-R1 for Django methods.
We present both cases where no original developer documentation existed and cases where the original docstring was available and incorporated.

\subsubsection{Examples Without Original Django Documentation}

\paragraph{Example 1: \texttt{Article.save}}\mbox{}\\[-0.5\baselineskip]

Method declaration:
\begin{lstlisting}[style=prompt]
def save(self, *args, **kwargs):
\end{lstlisting}

Method body:
\begin{lstlisting}[style=prompt]
if not self.id:
    self.created = datetime.date.today()
return super().save(*args, **kwargs)
\end{lstlisting}

Generated documentation (DeepSeek):
\begin{lstlisting}[style=prompt]
* What it does
Sets the 'created' date to the current date if the instance is new (when id is not set).

* Main inputs
*args and **kwargs passed to parent save method.

* Output / side effects
Saves the model instance to database and updates 'created' field for new instances.
\end{lstlisting}

\pagebreak
\paragraph{Example 2: \texttt{CustomErrorMessage.clean}}\mbox{}\\[-0.5\baselineskip]

Method declaration:
\begin{lstlisting}[style=prompt]
def clean(self):
\end{lstlisting}

Method body:
\begin{lstlisting}[style=prompt]
if self.name1 == "FORBIDDEN_VALUE":
    raise ValidationError(
        {"name1": [ValidationError("Model.clean() error messages.")]}
    )
elif self.name1 == "FORBIDDEN_VALUE2":
    raise ValidationError(
        {"name1": "Model.clean() error messages (simpler syntax)."}
    )
elif self.name1 == "GLOBAL_ERROR":
    raise ValidationError("Global error message.")
\end{lstlisting}

Generated documentation (DeepSeek):
\begin{lstlisting}[style=prompt]
* What it does
Validates the model instance by checking if 'name1' contains forbidden values.
Raises field-specific validation errors for "FORBIDDEN_VALUE" (detailed error
structure) and "FORBIDDEN_VALUE2" (simplified syntax), or a global validation
error for "GLOBAL_ERROR".

* Main inputs
Current value of the model's 'name1' field during validation.

* Output / side effects
Raises ValidationError with structured error messages attached to the 'name1'
field or as global errors, blocking model save until resolved.
\end{lstlisting}

\subsubsection{Examples With Original Django Documentation}

\paragraph{Example 1: \texttt{StartsWithRelation.field} }\mbox{}\\[-0.5\baselineskip]

Method declaration with original Django docstring:
\begin{lstlisting}[style=prompt]
def field(self):
    """
    Makes ReverseManyToOneDescriptor work in both directions.
    """
\end{lstlisting}

Method body:
\begin{lstlisting}[style=prompt]
return self.remote_field
\end{lstlisting}

Generated documentation (DeepSeek):
\begin{lstlisting}[style=prompt]
* What it does
Provides access to remote_field property to enable bidirectional
ReverseManyToOneDescriptor.

* Main inputs
None (property accessor).

* Output / side effects
Returns remote_field reference enabling descriptor functionality in both
directions.
\end{lstlisting}

\paragraph{Example 2: \texttt{Article.articles\_from\_same\_day\_2}}\mbox{}\\[-0.5\baselineskip]

Method declaration with original Django docstring:
\begin{lstlisting}[style=prompt]
def articles_from_same_day_2(self):
    """
    Verbose version of get_articles_from_same_day_1, which does a custom
    database query for the sake of demonstration.
    """
\end{lstlisting}

Method body:
\begin{lstlisting}[style=prompt]
from django.db import connection

with connection.cursor() as cursor:
    cursor.execute(
        """
        SELECT id, headline, pub_date
        FROM custom_methods_article
        WHERE pub_date = %s
            AND id != %s""",
        [connection.ops.adapt_datefield_value(self.pub_date), self.id],
    )
    return [self.__class__(*row) for row in cursor.fetchall()]
\end{lstlisting}

Generated documentation (DeepSeek):
\begin{lstlisting}[style=prompt]
* What it does
Retrieves same-day articles using direct database query.

* Main inputs
self (instance).

* Output / side effects
Returns list of Article instances created from raw query results.
\end{lstlisting}

\twocolumn

\twocolumn

\end{document}